\documentclass[journal]{IEEEtran}
\ifCLASSINFOpdf
  \usepackage[pdftex]{graphicx}
\else
  \usepackage[dvips]{graphicx}
\fi
\usepackage{amsmath}
\usepackage{booktabs}
\usepackage{amsmath,amssymb,amsfonts}
\usepackage{subfigure}

\usepackage{acronym}
\acrodef{DSM}{decision support model}
\acrodef{MCT}{minimum connecting time}
\acrodef{GMM}{Gaussian mixture model}
\acrodef{ML}{machine learning}
\acrodef{EDA}{exploratory data analysis}
\acrodef{AUC}{area under the curve}
\acrodef{ROC}{receiver operating characteristic}
\acrodef{PR}{precision-recall}
\acrodef{TP}{true positives}
\acrodef{FP}{false positives}
\acrodef{TN}{true negatives}
\acrodef{FN}{false negatives}
\acrodef{TPR}{true positive rate}
\acrodef{FPR}{false positive rate}
\acrodef{SS}{Schengen to Schengen}
\acrodef{SN}{Schengen to Non-Schengen}
\acrodef{NS}{Non-Schengen to Schengen}
\acrodef{NN}{Non-Schengen to Non-Schengen}
\acrodef{OR}{Operations Research}

\begin{document}
%
\title{Decision Support Models for Predicting and Explaining Airport Passenger Connectivity from Data}
%
%
%

\author{Marta~Guimarães, 
        Cláudia~Soares, 
        and~Rodrigo~Ventura,
\thanks{M. Guimaraes and R. Ventura are with Instituto Superior Técnico, Universidade de Lisboa, Portugal. e-mail: \texttt{\{marta.guimaraes,rodrigo.ventura\}@tecnico.ulisboa.pt}}
\thanks{C. Soares is with NOVA LINCS, Computer Science Department, NOVA School of Science and Technology, Universidade NOVA de Lisboa, 2829-516 Caparica, Portugal. (e-mail: \texttt{claudia.soares@fct.unl.pt}).}
}

%
%

\markboth{Journal of \LaTeX\ Class Files,~Vol.~14, No.~8, August~2015}%
{Shell \MakeLowercase{\textit{et al.}}: Predicting and Explaining Airport Passenger Connectivity}
%



\maketitle

\begin{abstract}
Predicting if passengers in a connecting flight will lose their connection is paramount for airline profitability.
We present novel machine learning-based decision support models for the different stages of connection flight management, namely for strategic, pre-tactical, tactical and post-operations.
We predict missed flight connections in an airline's hub airport using historical data on flights and passengers, and analyse the factors that contribute additively to the predicted outcome for each decision horizon. Our data is high-dimensional, heterogeneous, imbalanced and noisy, and does not inform about passenger arrival/departure transit time.
We employ probabilistic encoding of categorical classes, data balancing with Gaussian Mixture Models, and boosting. 
For all planning horizons, our models attain an \ac{AUC} of the \ac{ROC} higher than 0.93. SHAP value explanations of our models indicate that scheduled/perceived connection times contribute the most to the prediction, followed by passenger age and whether border controls are required.
\end{abstract}

\begin{IEEEkeywords}
Airline schedule planning, Model explanations, Data-driven operations, Imbalanced classification, Decision support models.
\end{IEEEkeywords}

%
\IEEEpeerreviewmaketitle

\section{Introduction}
\label{sec:introduction}

\begin{figure*}[h!]
    \centering
    \includegraphics[width=0.8\linewidth]{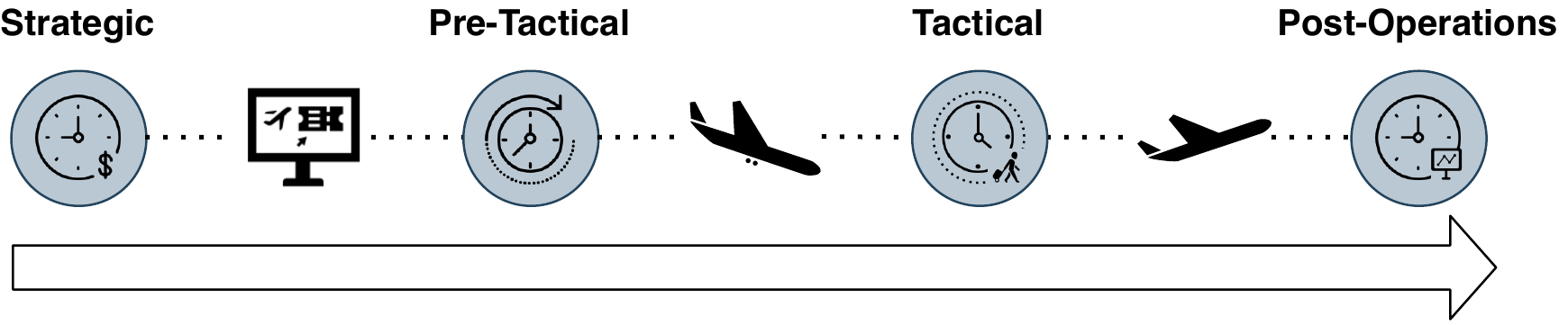}
    \caption{Proposed Decision Support Models. The first model that appears in the timeline of events is the Strategic model. It considers a long-term strategic approach. Once the flights schedule is made publicly accessible and passengers purchase their flights, new information becomes available and the Pre-Tactical model can be defined. After passengers disembark there is no information about their location at the airport. Thus, while the previous models worked in the long or short term, the Tactical model intends to support decisions occurring in real time at the airport. The last proposed model, Post-Operations, works as an analysis model and not a predictive one. Since at this stage all the information regarding the departure is known, it can be used to evaluate through historical data whether it is possible to make decisions for similar future events.}
    \label{fig:models}
\end{figure*}

We formulate the problem of predicting passenger connectivity in an airline's hub airport from flight and passenger data, which has seldom been studied. It is framed as an imbalanced classification problem where a \ac{ML} model is used to predict whether a passenger misses the connection or not. The proposed solution consists of \acp{DSM} that can be employed at different stages of connecting flights management, given the information available at each context. Our cost assessment indicates high cost reductions on our data by using preventive measures, contrasting with reactive costs.
Our models further identify key factors that impact passengers’ connection times.
The proposed solution is novel, by (1) integrating the \ac{ML} pipeline with airline operations and management, by (2) evidencing the relevant variables for accurate prediction in a real-world dataset, and by (3) considering the impact of the \ac{ML} pipeline in the operation cost.

One of the most challenging problems for airlines is schedule planning. The airline schedule planning problem largely defines the market share that an airline captures, and hence it is a key factor of the airline's profitability. Nowadays, an airline has to manage hundreds of flights per day; multiple aircraft; hundreds of airports (hubs and spokes); constraints regarding air traffic control, airport slots, gates, crew and passengers; while dealing with complex issues such as pricing and competing airlines. Therefore, designing a flight schedule that maximizes airline's revenues clearly is an extraordinarily complex problem~\cite{Barnhart2004AirlineOpportunities}. 
 
This type of schedule planning is based on the fact that the time at the airport is adjusted to what is strictly necessary to make the connection between flights and that the passenger is able to make the necessary circuit between landing/boarding.
 

A better planning of connecting flights is directly related to airline's profitability since an airline’s demand is influenced by: i) schedule of competing airlines, ii) delays and missed connections and iii) passenger goodwill.

To plan operational schedules, airlines today engage in a complex decision-making process, that becomes intractable due to mutual dependencies between airlines flight schedules, and the high-dimension of the considered variables.

Scheduling optimization techniques from the \ac{OR} community~\cite{Etschmaier1985AIRLINEOVERVIEW.} demand modeling the overall complex system of passenger flows, airline competing schedules, and random perturbations. As problems grow in size, and execution times become more demanding, \ac{OR} techniques are not as useful for the dispatch stages of pre-tactical and tactical management, as for the long-term strategic scenario. Accountability of systems is also a problem: in real-life operations, it is not possible to use automatic scheduling systems without human supervision, because of responsibility attribution for the choices made. Our approach is, thus, not to perform automatic scheduling, but to provide predictive support to a human decision-maker.

Moreover, as our goal is to take into account passenger features in the pre-tactical and tactical stages, \ac{ML}
approaches are the solution of choice.

Our work capitalizes on data science procedures over massive datasets collected by airlines, in order to produce a novel \ac{DSM} that can aid flight connection management for all stages, from strategic planning to dispatch.

We use decision intelligence to help decision-makers to optimize the flight schedule planning and to establish the \ac{MCT}. Decision intelligence uses data science to take advantage of the large amounts of available data to build effective~\acp{DSM}.

\subsection{Related Work}

There is a vast literature on the broader topic of scheduling flights, coupled or not with maintenance, fleet management, and airport capacity.

\paragraph*{Scheduling problems}
Some previous work have extensively studied optimization approaches for fleet assignment, aircraft maintenance routing and crew scheduling \cite{Barnhart2004AirlineOpportunities, Lohatepanont2001AirlineAlgorithms}. These are, in fact, important problems in \ac{OR}, with a vast and interesting literature. 
The problem of defining a flight schedule that aims to promote a robust operating system has been studied by some researchers, like Lan et al.~\cite{Lan2006PlanningDisruptions}, which presented two new approaches to minimize passenger disruptions and achieve robust airline schedule plans. 
Recently, Wu et al.~\cite{Wu2017AClosure} developed a fast solving method to large airline disruption problems specifically caused by airport closure. 
A previous paper~\cite{Burke2010AScheduling} investigates simultaneous flight re-timing and aircraft rerouting, subject to a fixed fleet assignment.
A number of papers have addressed the flight scheduling and fleet assignment simultaneously, including \cite{Lohatepanont2004AirlineAssignment, Yan2008AnDemands, Pita2013IntegratedCongestion}.
Yan et al.~\cite{Yan2007CoordinatedAirlines} introduced coordinated scheduling models for allied airlines, where the scheduling problem was enriched by the nuanced constraints of airline alliances. 
Authors in~\cite{Aloulou2010RobustRetiming} addressed the challenging issue of improving schedules robustness without increasing the planned costs. Robust scheduling is used to mitigate the impact of unforeseeable disruptions on profits, but can be too conservative in practice. Atkinson and coauthors~\cite{Atkinson2016RobustInefficiencies} examined how effectively practices such as flexibility to swap aircraft, flexibility to reassign gates, and scheduled aircraft downtime accomplish this goal.
A new algorithm to accurately calculate and minimize the cost of propagated delay was introduced by~\cite{Dunbar2014AnRe-timing}, in a framework that integrates aircraft routing and crew pairing.

\paragraph*{Airline frequency competition}
Airline frequency competition is partially responsible for the growing demand for airport resources. \cite{Vaze2012ModelingMitigation} proposed a model for airline frequency competition under slot constraints, looking for Nash equilibria in a competitive game, addressed by dynamic programming techniques.
\cite{Jiang2009DynamicScheduling} developed a dynamic
scheduling approach that aims to match capacity to demand given the many operational constraints that restrict possible assignments.

\paragraph*{Data-based research}
Flight schedules are highly sensitive to delays and witness these events on a very frequent basis.
The authors in~\cite{Mueller2002AnalysisCharacteristics} analyse departure and arrival data for ten major airports in the United States and characterize the delay distributions for traffic forecasting algorithms.
The airport-level causal relationship between delays, delay propagation, and delay causes was investigated using Bayesian networks~\cite{Xu2004EstimationNetworks}. Bayesian networks were also used to account for multiple connecting sources (aircraft, cabin crew, and pilots) and passenger connections~\cite{Wu2019ModellingModel}.
In a recent article, Guleria et al.~\cite{Guleria2019AFlights} developed a multi-agent-based method to predict the reactionary delays of flights, given the magnitude of primary delay that the flights witness at the beginning of the itinerary.
The authors in~\cite{Zhong2019ApplicationRecognition} proposed an application of Non-Negative Tensor Factorization for airport flight delay pattern recognition. This interesting work finds patterns in real-valued variables related with flights, while in our problem, we consider passenger features, many of them of ordinal or categorical type.
To understand the mechanism of delay propagation from the perspective of multiple airports,~\cite{Xiao2020StudyEntropy} proposed a low-dimensional approximation of conditional mutual information for transfer entropy.
Chen et al.~\cite{Chen2020VFDP:Map} designed a visual analysis system for studying the delay propagation trends in one region and representing the relationship of delays occurring in multiple airports.
Güvercin et al.~\cite{Guvercin2021ForecastingNetworks} proposed a Clustered 
Airport Modelling approach to address the problem of forecasting flight delays of an airport. This work used the network information as well as the delay patterns of similar airports in the network. 
The authors in~\cite{Cai2021AGraphs} investigated the delay prediction problem from a network perspective by developing an approach based on the graph convolutional neural network.
Efthymiou et al.~\cite{Efthymiou2019TheAirport} performed an empirical analysis focusing on the impact of delays on customers’ experience and satisfaction.
Some researchers approach the specific problem of predicting flight delays from flight information only. \cite{Khanmohammadi2016AAirport} introduce a new multilevel input layer artificial neural network and \cite{Belcastro2016UsingDelays} implemented a parallel version of the Random Forest data-classification algorithm for predicting flight delays.

Besides flight delays, another important component in flight scheduling is the flight
block time and its reliability.
\cite{Tian2020AssessmentStudy} proposes a methodology for evaluating the flight block time under different delay time windows.

\paragraph*{Methods using passenger information}
Typically, passenger data is not publicly available making it difficult to explore passenger-centric problems.
Bratu et al.~\cite{Bratu2005AnData} developed a passenger delay calculator to compute passenger delays and to establish relationships between passenger delays and cancellation rates, flight leg delay distributions, load factors, and flight schedule design.
Later, the same authors proposed in~\cite{Bratu2006FlightRecovery} airline recovery decision models that select flight leg departure times and cancellations that, like conventional models, minimize operating costs, but are extended to include the resulting delay and disruption costs experienced by passengers. 
The recent paper~\cite{Rosenow2020EvaluationCosts} performed an evaluation of strategies to reduce the cost impacts of flight delays on total network costs. The inclusion of individual transfer passengers in the delay cost balance of an airline involves the uncertainty of the number of these passengers. Previously, Barnhart et al.~\cite{Barnhart2014ModelingSystem} developed a multinomial logit model for estimating historical passenger travel and extend a previously developed greedy re-accommodation heuristic for estimation of the resulting passenger delays. 
A recent approach to study passengers’ transfer journeys using flight and passenger datasets~\cite{Guo2020ForecastingLearning}, does provide decision support in real-time. It uses customized machine learning algorithms to predict passengers' connection times. 
This approach differs from the one proposed in this work, since it has access to information about historical connection times, which allows to make a regression model. Our problem only allows for knowing if the connection was successful or not.

\paragraph*{Contributions}
We propose a novel set of \acp{DSM} with similar, recognizable structure, for connecting flights schedule planning in different horizons: strategic (Section~\ref{sec:strategic}), pre-tactical (Section~\ref{sec:pretactical}), tactical (Section~\ref{sec:tactical}), and, as an analysis tool, a post-operations model (Section~\ref{sec:postoperations}), without requiring the connection time of each passenger.
Our approach is data-driven and, thus, assumes that the data distribution remains similar to the data used for training the models. Nevertheless, such assumption is valid in most periods of time. Under this assumption, we exhibit AUC of the ROC of 0.98 for the long-term \ac{DSM} and pre-tactical, and 0.99 for tactical (Section~\ref{sec:results}). Our operational cost analysis shows that, for our data, the use of the proposed set of \acp{DSM} for decision-making leads to high cost reductions (Section~\ref{sec:operational-costs}). We investigated SHAP-based explainations of our models, for sanity-check of the models and to extract knowledge about the interactions of flights and passengers.

\section{Data Analysis and Preparation}
\label{sec:EDA}

TAP Air Portugal's operating model is based on the hub and spoke philosophy, with its operation concentrated at Lisbon airport, where it allows passenger flows to connect between flights that land and take off at Humberto Delgado Airport. This is one of the most important European gateways to Brazil and Africa and the biggest European airport serving South America among the Star Alliance hubs. In 2019, Lisbon airport handled 31.2 million commercial passengers. From these, 17.1 million were TAP passengers~\cite{2019RelatorioConsolidadas}.

Historical data for all of 2019 was provided by TAP. The data consists of three different sets: i) SEF (border control), ii) Hub (flight) and iii) Pax (passenger) datasets. 
The Hub dataset includes all the information about the departure and arrival flights such as the scheduled arrival/departure time, gate number, etc. This dataset contains 108 features and 345035 samples. From these samples, 174030 (50.4\%) are related with the arrivals and 171005 (49.6\%) with the departures.
The Pax (short for passenger) dataset contains the passenger's available travel information, e.g., their travel class and connection status. Lastly, SEF (border control service) dataset associates each pair of arrival/departure airports with a binary variable that indicates if the passenger has to pass through the passport control or not. The SEF dataset itself does not provide much information, but associated with the Pax dataset may generate interesting results. For this reason, before proceeding with further analysis, a new column was added to the Pax data to include this information. This dataset consists of 5034919 samples (rows) and 17 features (columns).

Both the Hub and Pax datasets represent big volumes of data. We performed an \ac{EDA} to understand the data. The most important discoveries of the \ac{EDA} were:
\begin{itemize}
    \item The Pax dataset contains the passenger's travel information, so each row represents a passenger's connection. If a passenger misses the original connecting flight, it is highly likely that there is another sample associated with the new connection;
    \item In the Pax dataset, it was found a systematic error on the arrival flight dates. A careful data cleaning process was performed to recover these data.
\end{itemize}
The data integration process was done through a primary key to ensure that for a given flight in the Pax dataset there was only one possible match in the Hub dataset. After this merge, feature engineering was performed. Some of the most relevant new features included/transformed were:
\begin{itemize}
    \item \textbf{Perceived Connection Time.} This variable is the scheduled off-blocks time of the departure flight minus the actual on-blocks time of the arrival. It represents the information that the passengers have regarding the connecting time. Thus, may be an indicator of the stress level of the connecting passenger, which influences the speed with which a person moves along the airport \cite{Guo2020ForecastingLearning}.
    \item  \textbf{Traffic Network.} This feature is important for European airports. It indicates if the passenger is travelling from and to airports within the Schengen area. Thus, there are four types of connections (traffic networks): \ac{SS}, \ac{SN}, \ac{NS} and \ac{NN}. When exiting or entering the Schengen area, passengers have to go through border control procedures, while travelling within Schengen countries no border control is done.
\end{itemize}

With the feature engineering process a significant amount of information was extracted from the original data and added to the already existing set of features. Thus, at this point, the data needs to be selected or removed to ensure that the data values fall within the context and formulation of the problem and also that features are not strongly correlated. Therefore, all the samples which appeared to present errors or related to passengers who never made it to the airport were deleted. Only flights from TAP were selected. As the objective of this work is to analyse the original passengers' connecting flights, only these samples were considered. With this procedure the dataset consists of 3.592.004 samples.

Regarding the missing values, as values are randomly omitted from the dataset, applying listwise deletion does not introduce a bias in the data.\footnote{Listwise deletion is a method for handling missing data in which a sample is removed from the dataset if any single value in that sample is missing. Such samples do not represent a high percentage of the total dataset.}

This method is applied before training each model and hence it depends on the feature selection process. Section~\ref{sec:DSM} explains in more detail the features considered for each problem. With this method, between 0.60\% and 3.90\% of the samples were dropped in each problem, which means that the statistical power of the data remains essentially unaltered.

In this work, the standardization method was used to perform feature scaling and the categorical features were encoded with target encoding~\cite{Micci-Barreca2001AProblems}. These procedures were fitted only on the training set and then those parameters were used to transform the test set to avoid leakage.

\section{Decision Support Models}
\label{sec:DSM}

The proposed \acs{DSM} are designed to inform the airline whether the connecting flight is at risk of being lost or to perform a post-hoc analysis. Taking into account the information available at a given time and context, four \acs{DSM} were defined: strategic, pre-tactical, tactical and post-operations (Figure \ref{fig:models}). Each \ac{DSM} consists of a \ac{ML} model that takes as input the data available for decision making, and outputs the prediction that supports the decision. Hence, all of these \acs{DSM} have the same structure, differing only in the available information used as input data. 

\subsection{Strategic}
\label{sec:strategic}
The first model that appears in the timeline of events is the strategic \ac{DSM}. It considers a long-term strategic approach and, since it is the first to be defined, it has greater influence on future events. Currently, the scheduling process of connecting flights is made assuming the previously defined \ac{MCT} as the only limitation to this process. The Strategic \ac{DSM} intends to support this process, replacing the pre-established (and uninformed) restriction by the predictions of the model which consider all the available past information.

Concerning the flights features available at this stage, the arrival and departure flight designators, named `TP From' and `TP To' respectively, are used. Each flight designator is associated with an arrival and departure airport and implicitly has information about the time of the day it occurs, since there are no repeated flight designators on each day. 

In most cases, the arrival and departure flights happen on the same day, so using features related with both flight dates tends to be quite redundant. Despite the data cleaning stage, the arrival date feature still contains missing values and so, the departure flight information was used in order to avoid dealing with more missing data. Two features regarding the departure flights were selected: day of the week and day of the month. Such features are expected to be useful in the sense that both represent seasonalities.

As already stated, the traffic network is considered to be useful since it may have an effect on the time needed to go from the arrival boarding gate to the departure one, whenever passengers have to go through border control.

After the feature selection process described above and applying listwise deletion, the data consists of 3,570,280 data points and 208,828 (5.85\%) belong to the minority class (class of missed flights), thereby confirming that the problem is imbalanced.

\subsection{Pre-Tactical}
\label{sec:pretactical}
After defining the Strategic \ac{DSM}, the flights schedule is made publicly available and passengers purchase their flights. Thus, passenger information is now available, and new features are added to the information that was previously used in the Strategic model. The main goal of the Pre-Tactical \ac{DSM} is to understand if and how information about passengers allows the \ac{ML} model to refine its predictions. This \ac{DSM} can be very advantageous in situations where it is necessary to make adjustments to the initially defined connection time. It can also be a useful tool for the airline to be aware of the general profile of passengers making a certain connection and to define strategies in advance to prevent missed connecting flights. For example, allocating an employee to guide certain passengers on the route they need to take at the airport in order to prevent them from getting lost along the way; or change the passenger's seat on the arrival flight so that the passenger is closer to the door and can get out of the plane more quickly to avoid missing the departure flight. 

The new features available at this stage are the passenger's gender and age, whether travelling in group or not and the classes in which the passenger is travelling on the arrival and departure flights (e.g., business and economy classes). By adding this information, the objective is to understand if the \ac{ML} model can somehow associate this characteristics to form a profile of passengers with more or less likelihood to miss the connection. This will allow to take conclusions such as: ``Does a male passenger traveling with a group of friends is more likely to miss the connecting flight?".  

This model can also be seen as a short-term strategic approach, since its impact is important from a short term decision perspective.

The data used to train and test the model consists of 3,545,578 samples. The minority class (missed
flights) is composed of 204,338 (5.76\%) instances.

\subsection{Tactical}
\label{sec:tactical}
The Tactical \ac{DSM} supports decisions about events occurring in real time at the airport regarding connecting flights and passengers. That is, when the information known is the arrival and the scheduled departure times. Thus, while the previous models worked in the long or short term, this one is characterized by the influence it can have in real time events. 
After the passenger disembarks there is no information about the location of the passenger at the airport. It is not possible to know if the passenger is waiting in line for passport control or in the shopping area. Therefore, it becomes complicated to make decisions, for instance to understand if it is worth delaying the boarding of the departure flight and, consequently, delaying other procedures, to ensure that passengers do not miss their connecting flight, or if it is worth paying some compensations and not delaying the departure flight. In this type of situations, the Tactical \ac{DSM} might be used to predict the number of passengers in this situation thereby providing more information to support this decision.

In addition to these examples, this model can also be used to make the decisions mentioned in the Pre-Tactical \ac{DSM}, but with more updated information. In fact, this is the reason why the models are called Pre-Tactical and Tactical.

All arrival flight information is considered to be known. As possible delays in the departure flights are not always known \textit{a priori}, these are not taken into account at this stage. It is thus assumed that there is no new information available regarding the departures. The new information available at this stage is the arrival flight delay. Even so, it was considered that it would be more relevant to directly use the feature that represents the scheduled connection time discounted by the arrival delay, i.e., the perceived connection time. Thus, of the features already used in the \acp{DSM} described above, the scheduled connection time was replaced by the perceived one.

The dateset is reduced to 3,545,578 samples, wherein 204,338 (5.76\%) belong to the minority class.

\subsection{Post-Operations}
\label{sec:postoperations}
The last proposed model is the Post-Operations \ac{DSM}. At this stage, all the information regarding the departure is known, namely the delay of the flight. Taking into account that the events have already occurred, there is no need to make a decision on those same events. Therefore, unlike the other models described, this model works as an analysis model and not as a predictive one. This model can be useful to evaluate through historical data whether it is possible to make decisions for similar future events. 

As the connection time is defined to be the interval between `on-blocks' and `off-blocks', the boarding stage and the number of buses used to carry the passengers from the boarding gate to the platform are included in this time interval. Therefore, the boarding time interval and number of buses used in the departure flight were added to the already used feature set.

It was decided to consider the actual connection time feature instead of using the scheduled time and the arrival and departure delays separately. Hence, the time that the passengers had to make the connection is introduced directly into the \ac{ML} model, instead of being deducted through the other mentioned features. 

The input data consists of 3,451,979 samples. From these samples 5.51\% belong the positive class.


Table \ref{tab:FeatureSelection} summarizes the features available for each model.

\begin{table}[!htb]
\caption[Features considered for each problem.]{Features considered for each problem, considering causal constraints.}
  \renewcommand{\arraystretch}{1} 
  \centering
  \begin{tabular}{lcccc}
    \toprule
    Features &   Strat. & Pre-Tact. & Tact. & Post-Oper.\\
    \midrule
    TP From   & \checkmark & \checkmark & \checkmark & \checkmark\\
    TP To  &  \checkmark & \checkmark & \checkmark & \checkmark\\
    Traffic Network & \checkmark & \checkmark & \checkmark & \checkmark\\
    Dep. Day  & \checkmark & \checkmark & \checkmark & \checkmark\\
    Dep. Month Day & \checkmark & \checkmark & \checkmark & \checkmark\\
    Boarding Delta  &  &  &  & \checkmark\\
    N Bus  &  &  &  & \checkmark\\
    Sex &   & \checkmark & \checkmark & \checkmark\\
    Age &   & \checkmark & \checkmark & \checkmark\\
    Is Group &  & \checkmark & \checkmark & \checkmark\\
    Class From &  & \checkmark & \checkmark & \checkmark\\
    Class To &   & \checkmark & \checkmark & \checkmark\\
    \midrule
    Sch. Conn. Time & \checkmark & \checkmark &  & \\
    Perceived Conn. Time &  &  & \checkmark & \\
    Actual Conn. Time &  &  &  & \checkmark\\
    \bottomrule
  \end{tabular}
  \label{tab:FeatureSelection}
\end{table}

\subsection{Baselines}
 The baseline classifiers are inspired by the criterion used by TAP which uses the \ac{MCT} established by ANA - Aeroportos de Portugal, defined as 60 minutes. The baseline classifier only uses one feature, the connection time. 
A connection is predicted to be missed if the time between flights is smaller than the \ac{MCT}.

\begin{figure}[t!]
    \centering
        \subfigure[ROC curve of the Strategic baseline model.]{\includegraphics[width=0.35\textwidth]{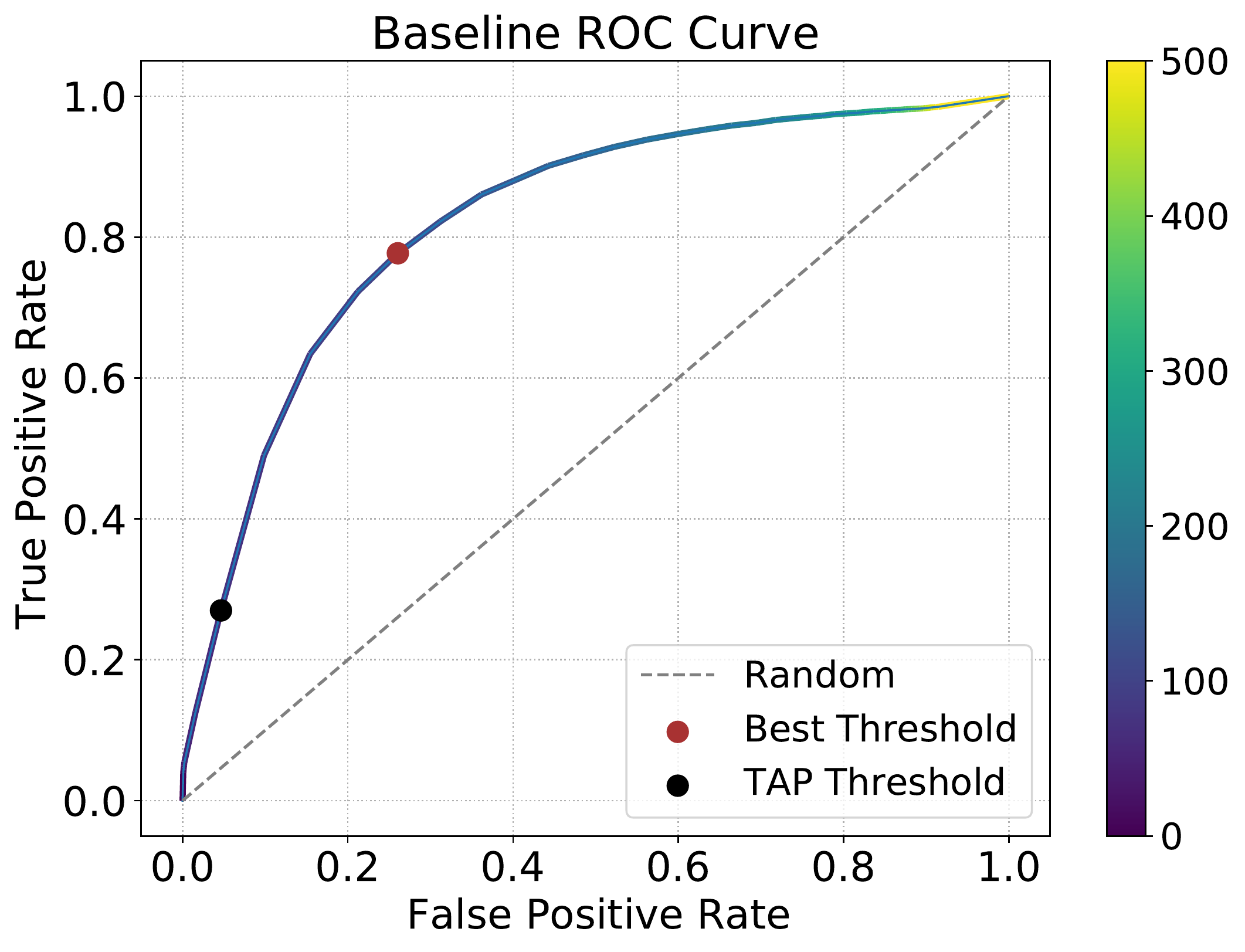}\label{fig:BaselineStrategicROC}}
        \subfigure[PR curve of the Strategic baseline model.]{\includegraphics[width=0.35\textwidth]{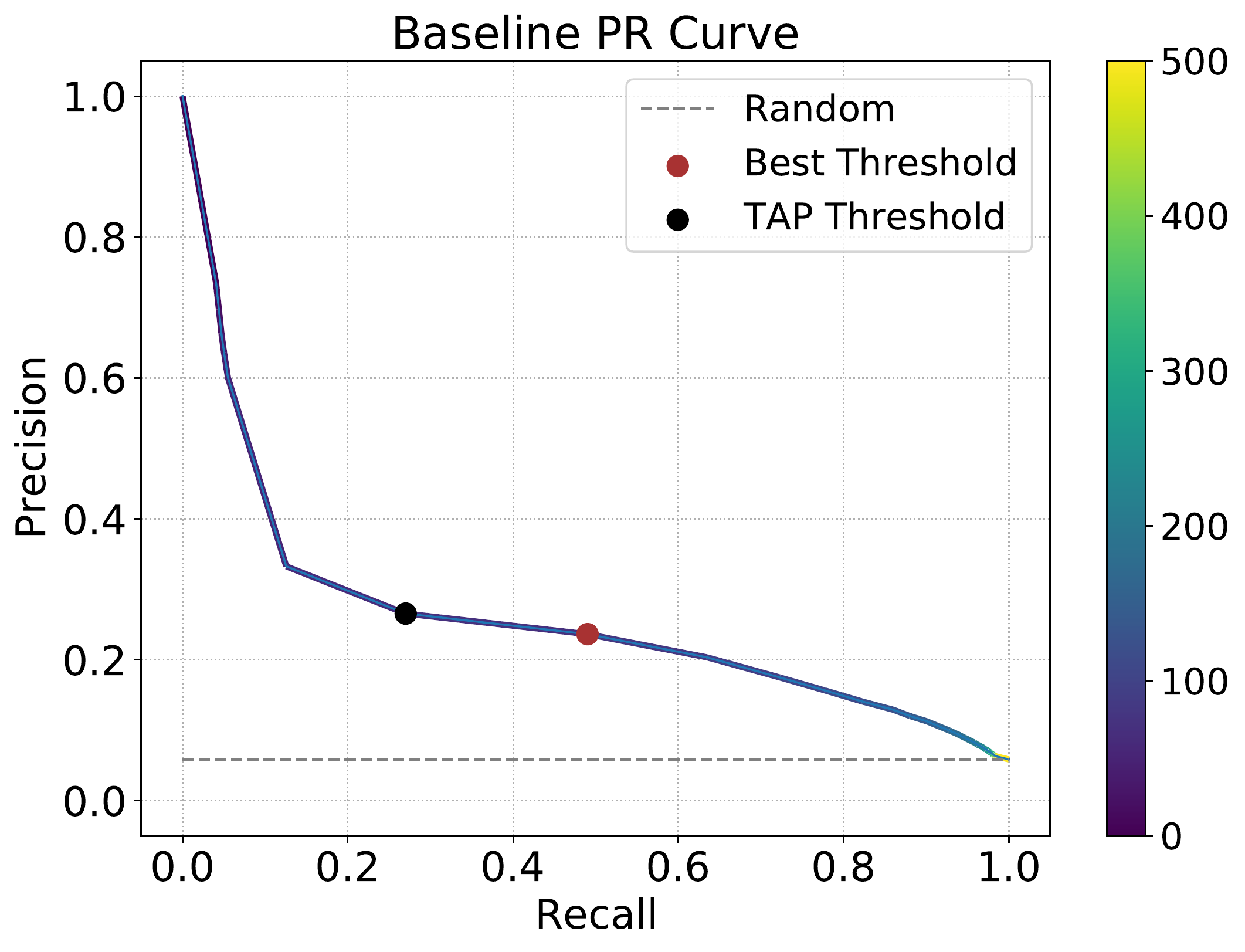}\label{fig:BaselineStrategicPR}}
    \caption{ROC and PR curves of the baseline model of the Strategic \ac{DSM}. We observe that a threshold model can be improved by data analysis only, with respect to the conventional threshold used nowadays. We will consider the optimal threshold. The colormap represents the threshold value in minutes. Further results are detailed in Section~\ref{sec:results}.}
    \label{fig:Baseline}
\end{figure}

Instead of fixing one threshold of connection time, each baseline classifier uses a range of values from 0 to 500 minutes with a 10 minutes step value, to select the one with the best performance (Figure \ref{fig:Baseline}). 

Depending on the problem, the connection time which the baseline classifier considers can be the scheduled, perceived time or actual.

\subsection{Model}

This is an imbalanced classification problem. A first step is to rebalance the training data so that the classification model is not biased by the sheer quantity of majority class examples. To this end, we will oversample the minority class of the training set. We use a principled probabilistic approach for learning the distribution of the minority class, the \ac{GMM}. Then, through a generative process, we will sample from the learned distribution.

The \ac{GMM} is a probabilistic model commonly used for both clustering and density estimation, that assumes that all data points are generated from a mixture of $K$ Gaussian distributions $\mathcal{N}(x ; \mu_k, \Sigma_k)$ of mean $\mu_k$ and covariance $\Sigma_k$,
\begin{equation}
    p(x) = \sum_{k =1} ^K \pi_k \mathcal{N}(x ; \mu_k, \Sigma_k),
\end{equation}
where $\pi_k$ are the mixture parameters, or prior probabilities of each model $k$, such that $\sum_{k=1}^K \pi_k = 1$.
This assumption is not a limitation because a \ac{GMM} $p(x)$ is known to approximate any probability distribution, if the number of models $K$ is sufficiently large~\cite{Sorenson1971RecursiveSumsb}, even for simple covariance structures, for example, considering $\Sigma_k$ as diagonal matrices. Gaussian mixtures allow for multimodality and can be fitted efficiently through the Expectation-Maximization algorithm.

We used the XGBoost classifier~\cite{Chen2016XGBoost:System} as it presents state of the art performance for classification tasks with tabular data, even under mild imbalance.

Determination of model hyperparameters for data augmentation and for classification is detailed in Section~\ref{sec:results}.

\section{Model training and results}
\label{sec:results}

\begin{figure*}[t!]
    \centering
        \subfigure[Strategic. ]{\includegraphics[width=0.2\textwidth]{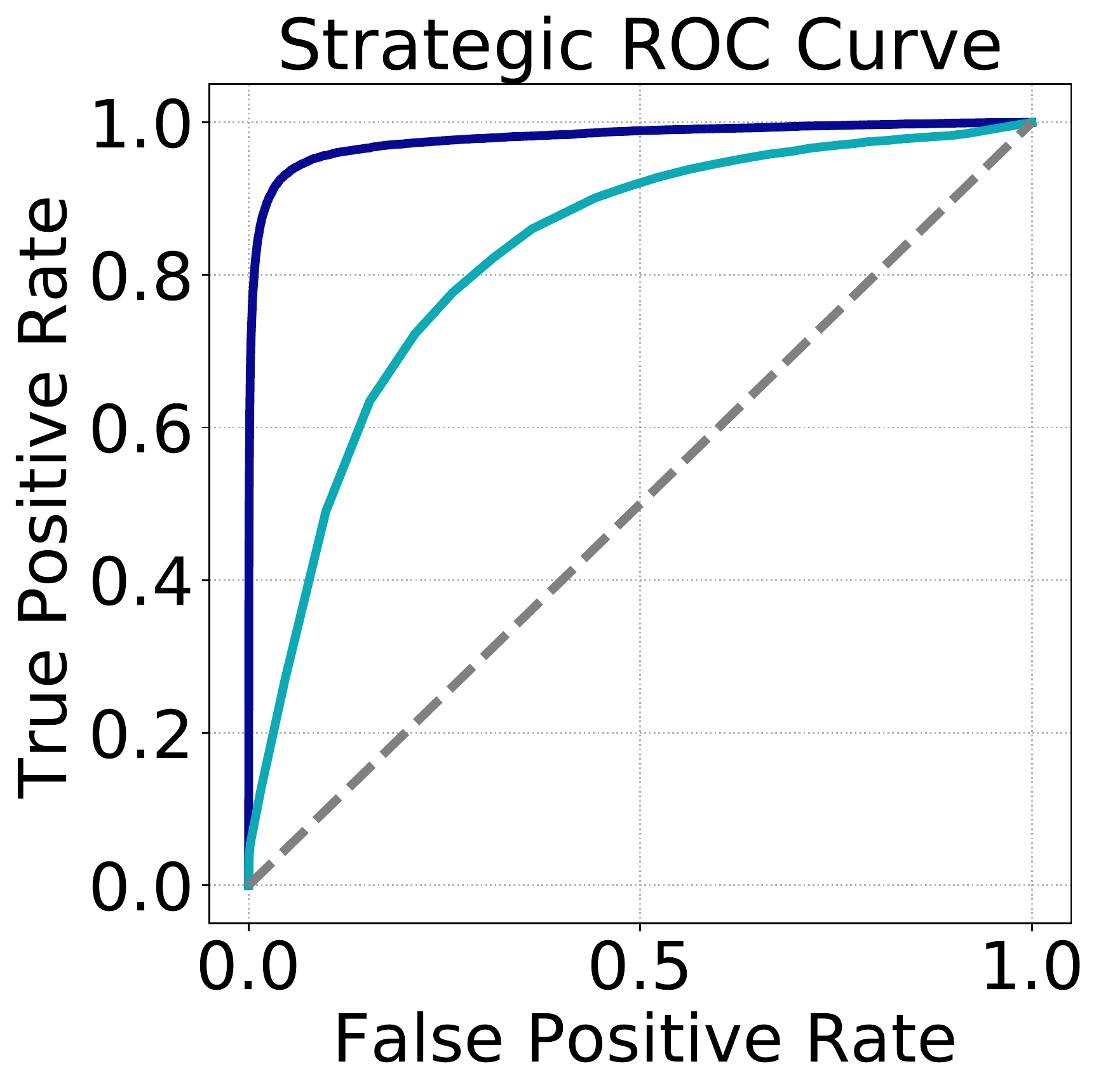}\label{fig:StrategicROC}}
        \subfigure[Pre-Tactical.]{\includegraphics[width=0.2\textwidth]{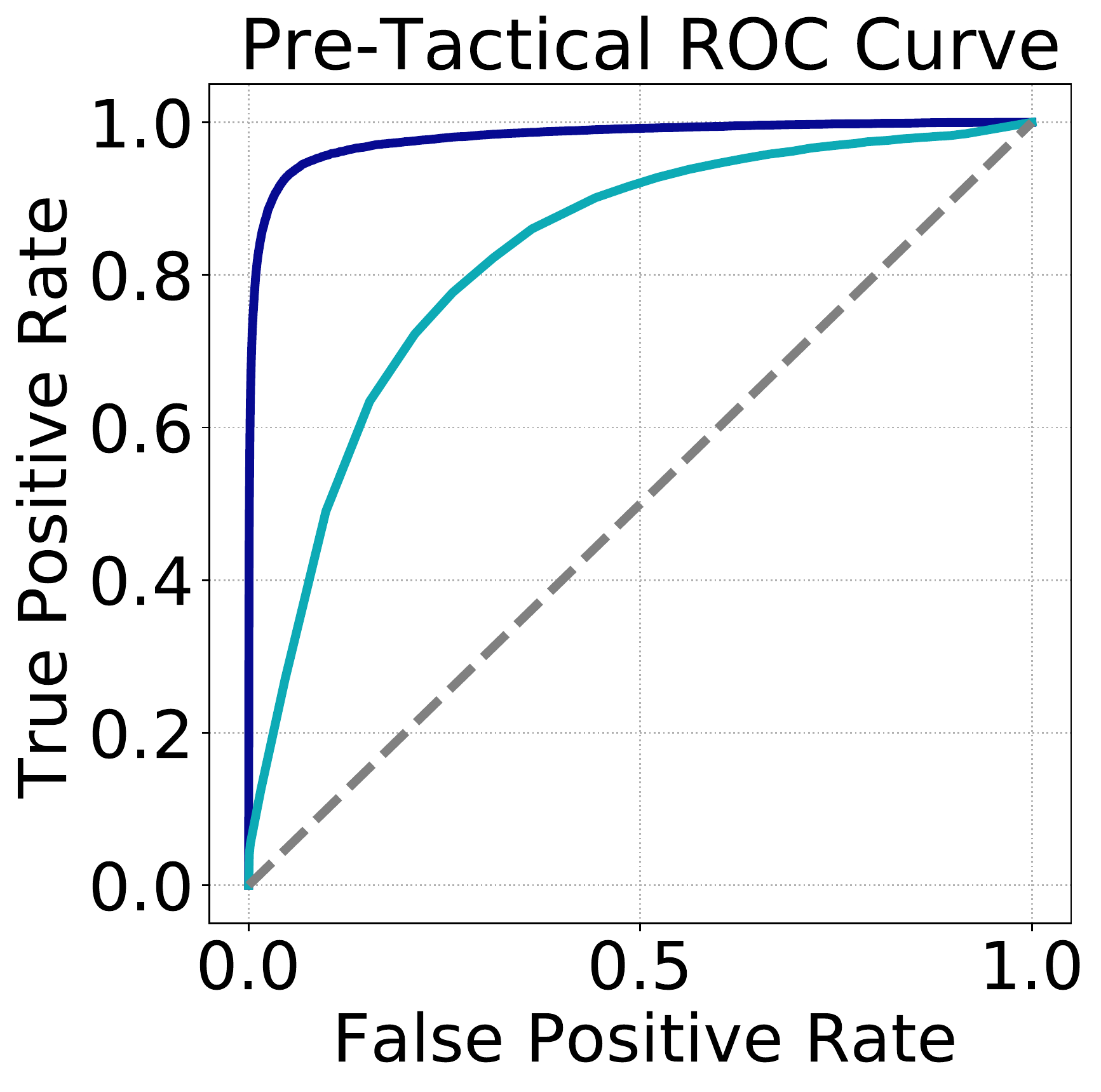}\label{fig:PreTacticalROC}}
        \subfigure[Tactical.]{\includegraphics[width=0.2\textwidth]{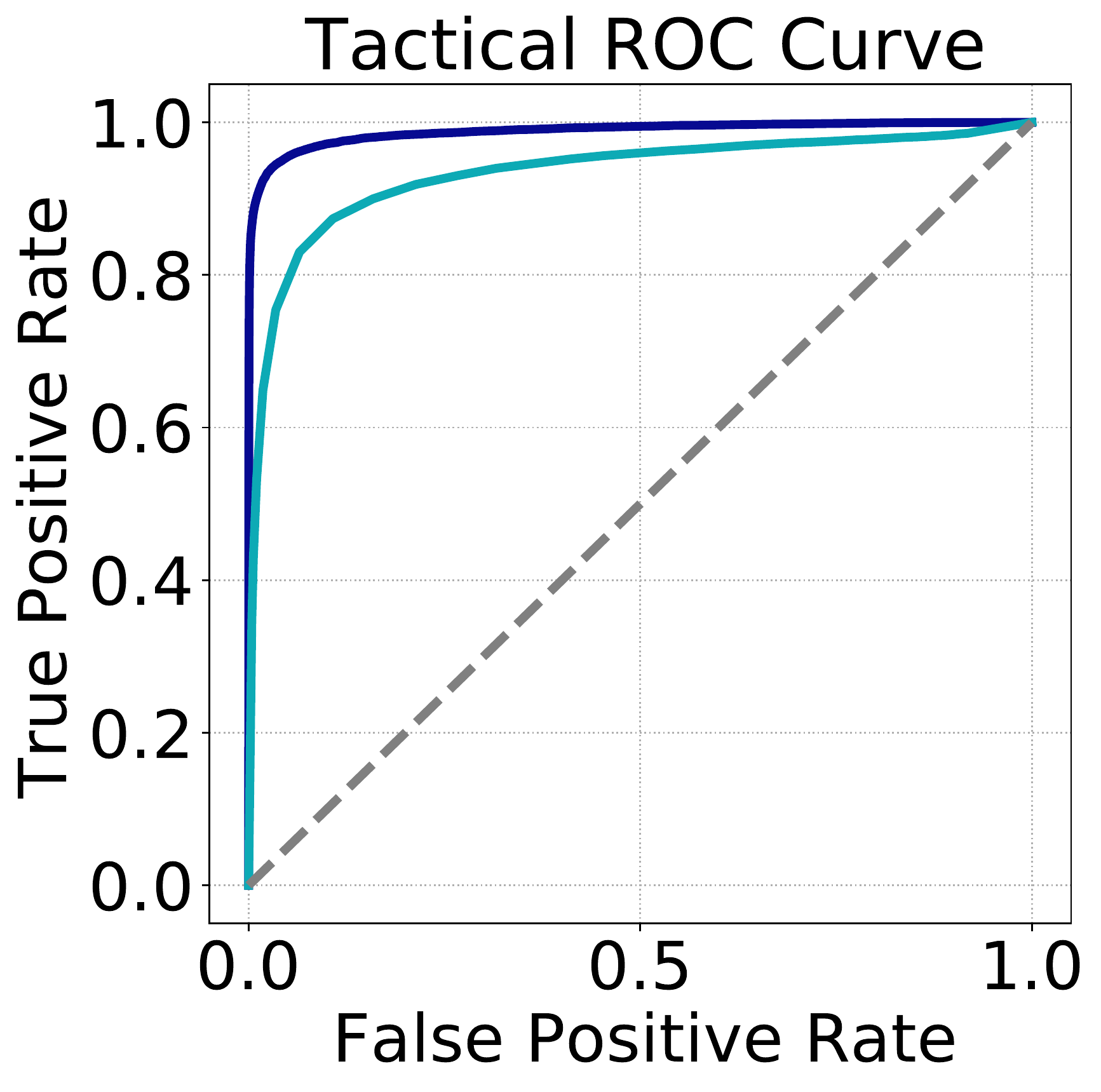}\label{fig:TacticalROC}}
        \subfigure[Post-Operations.]{\includegraphics[width=0.295\textwidth]{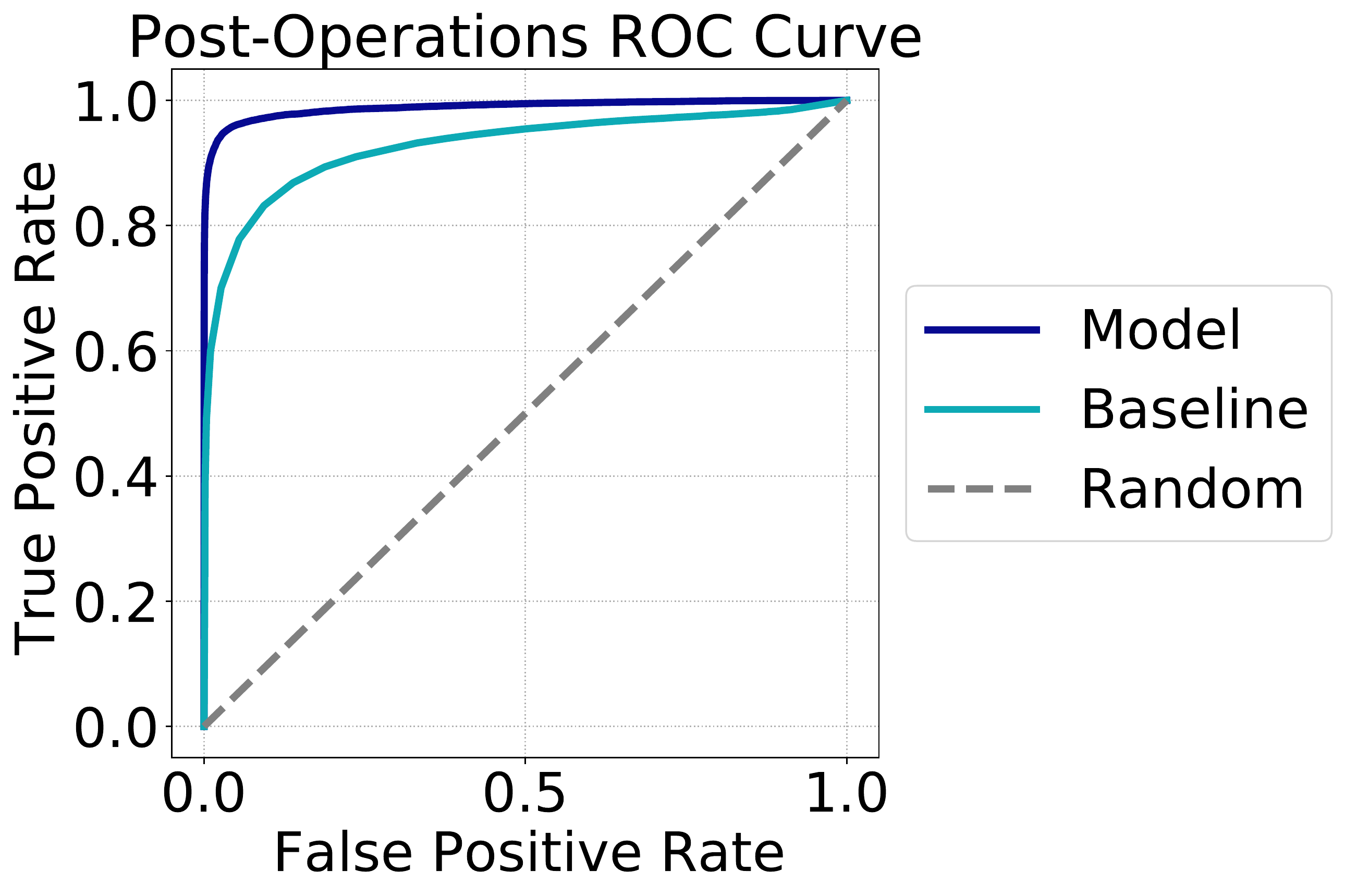}\label{fig:PostOperationsROC}}
    \caption{\ac{ROC} curves for the proposed models and the corresponding baselines for the test set. Our Machine Learning model always improves significantly over the baseline, and the improvement is greater in the predictive setting of strategic and pre-tactical planning frames.  The Strategic and Pre-Tactical Problems present an AUC$_\text{ROC}$ equal to 0.82 for the baseline solution and AUC$_\text{ROC}=0.98$ for both of the proposed models. The Tactical DSM yields an AUC$_\text{ROC}$ of 0.99 while for the corresponding baseline the AUC$_\text{ROC}$ is 0.93. The obtained values for the Post-Operations Problem were 0.93 and 0.99 for the baseline and the proposed DSM, respectively. }
    \label{fig:ResultsROC}
\end{figure*}
\begin{figure*}[t!]
    \centering
        \subfigure[Strategic.]{\includegraphics[width=0.2\textwidth]{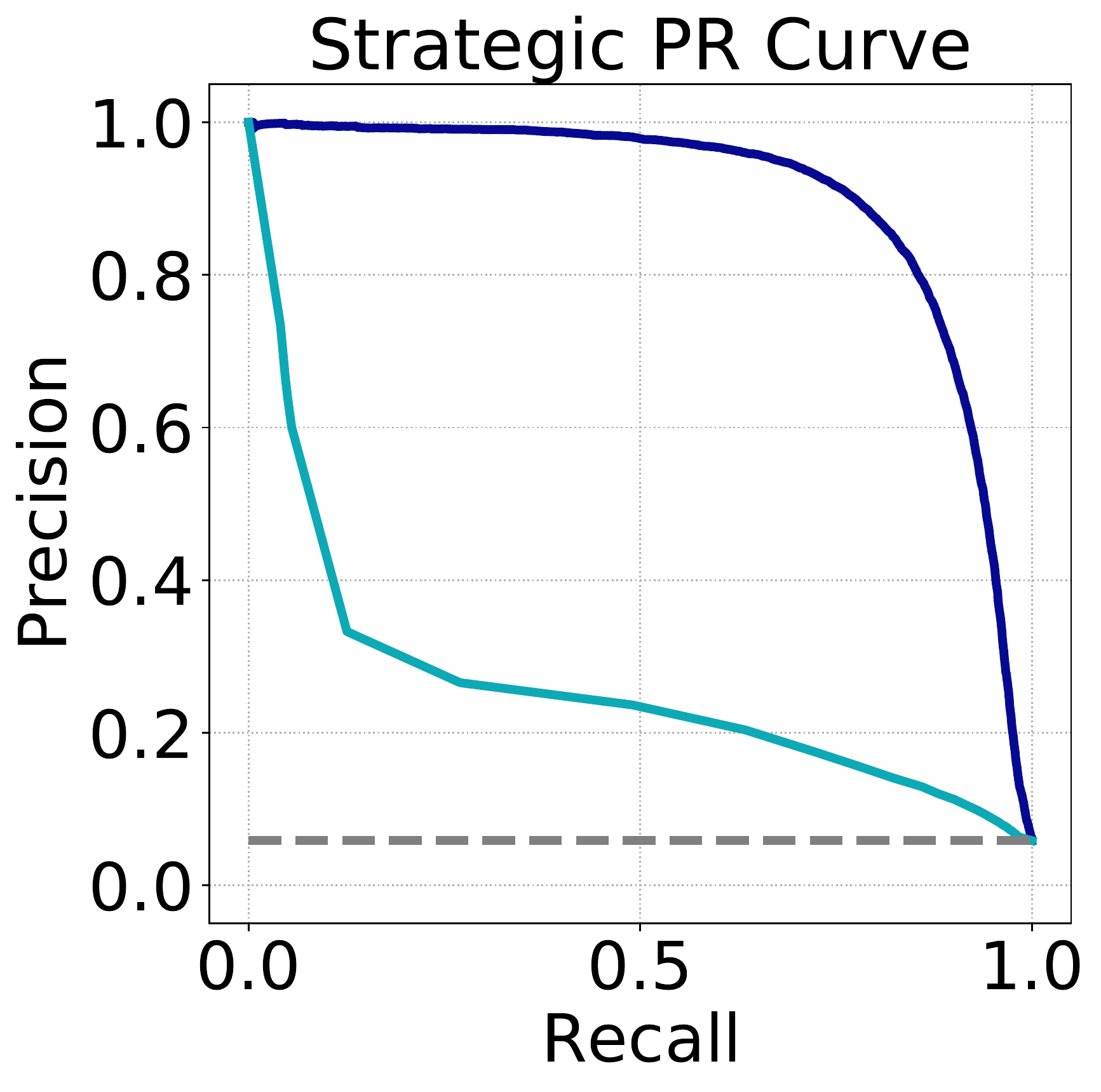}\label{fig:StrategicPR}}
        \subfigure[Pre-Tactical.]{\includegraphics[width=0.2\textwidth]{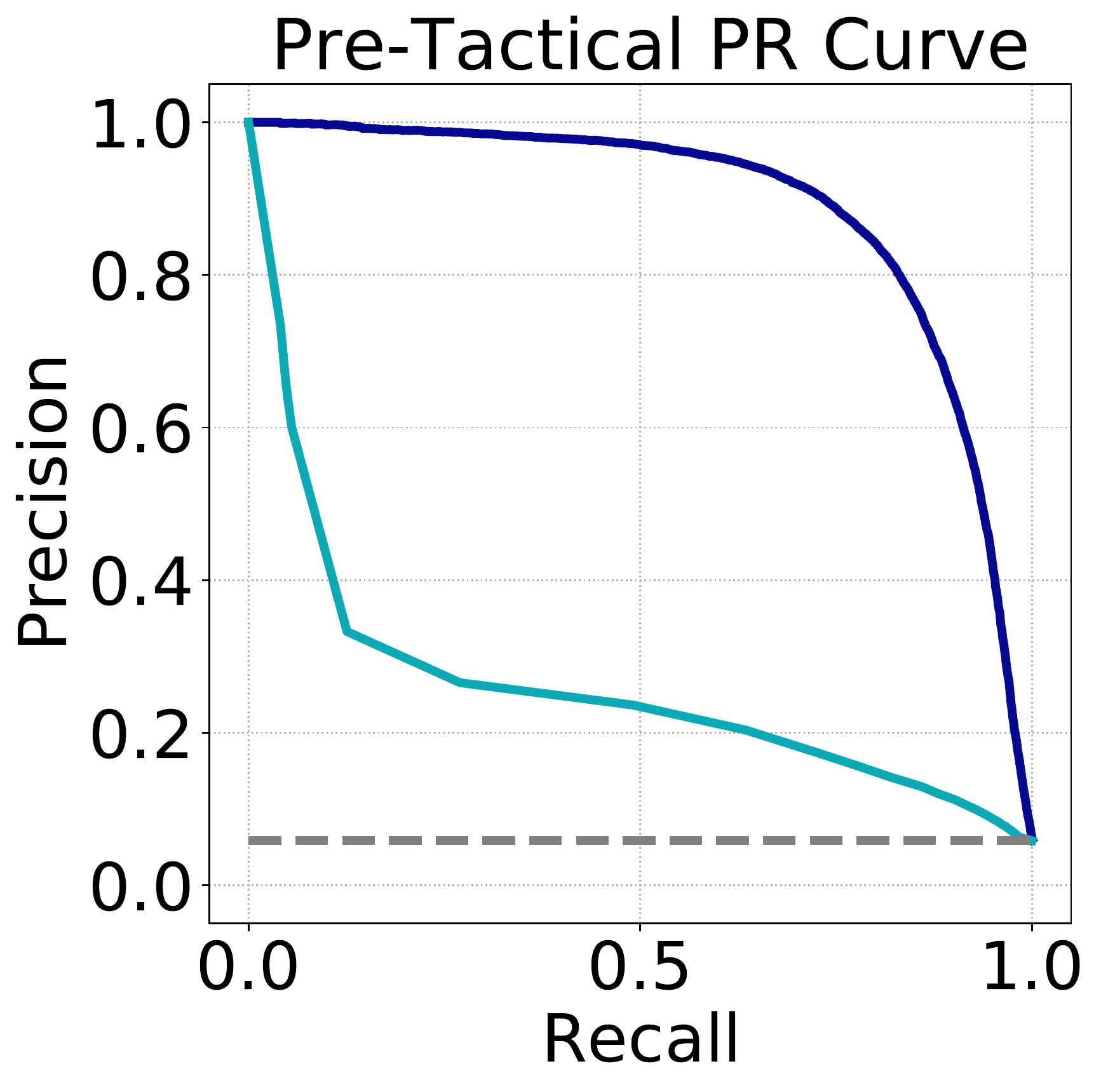}\label{fig:PreTacticalPR}}
        \subfigure[Tactical.]{\includegraphics[width=0.2\textwidth]{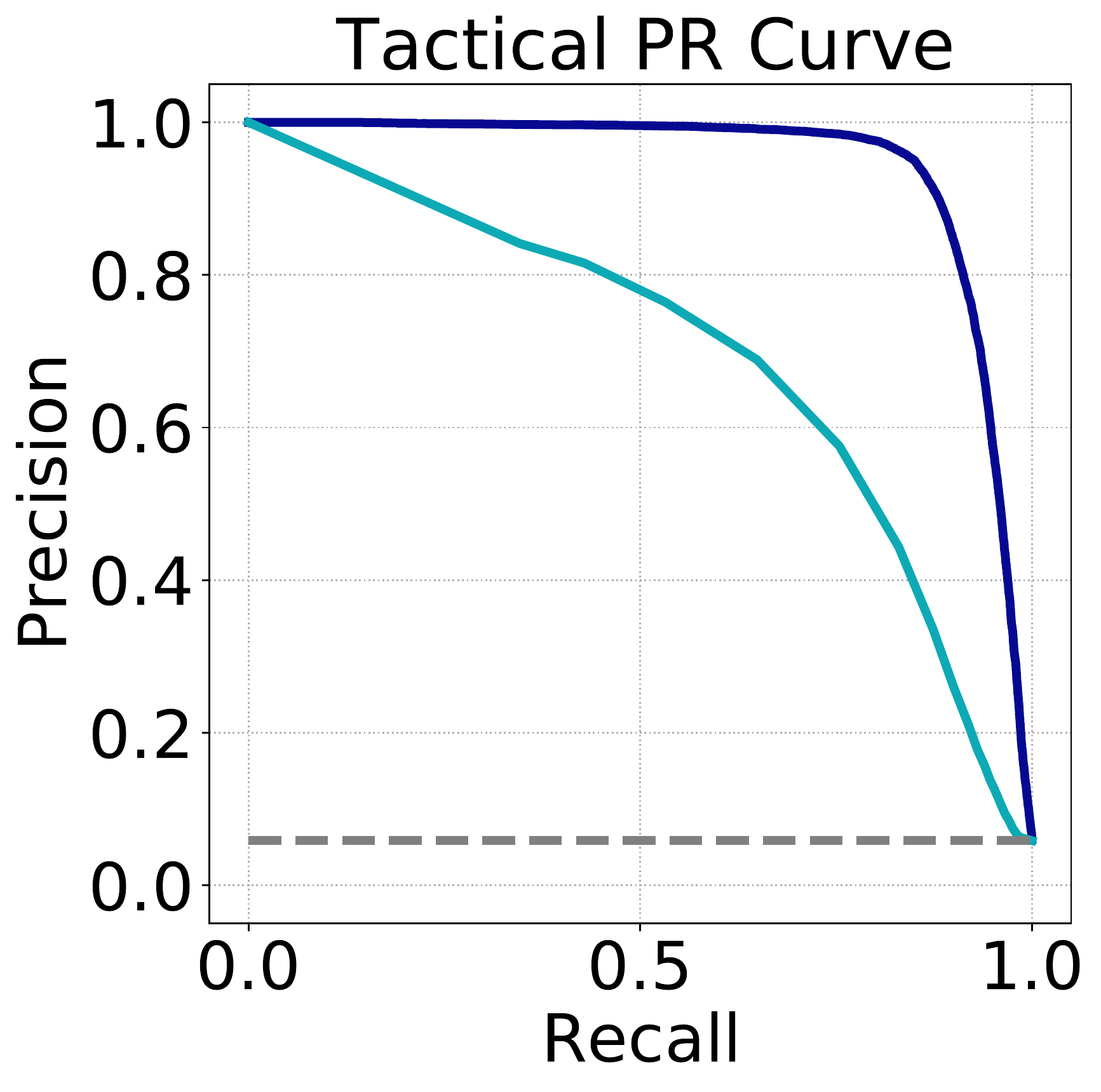}\label{fig:TacticalPR}}
        \subfigure[Post-Operations.]{\includegraphics[width=0.295\textwidth]{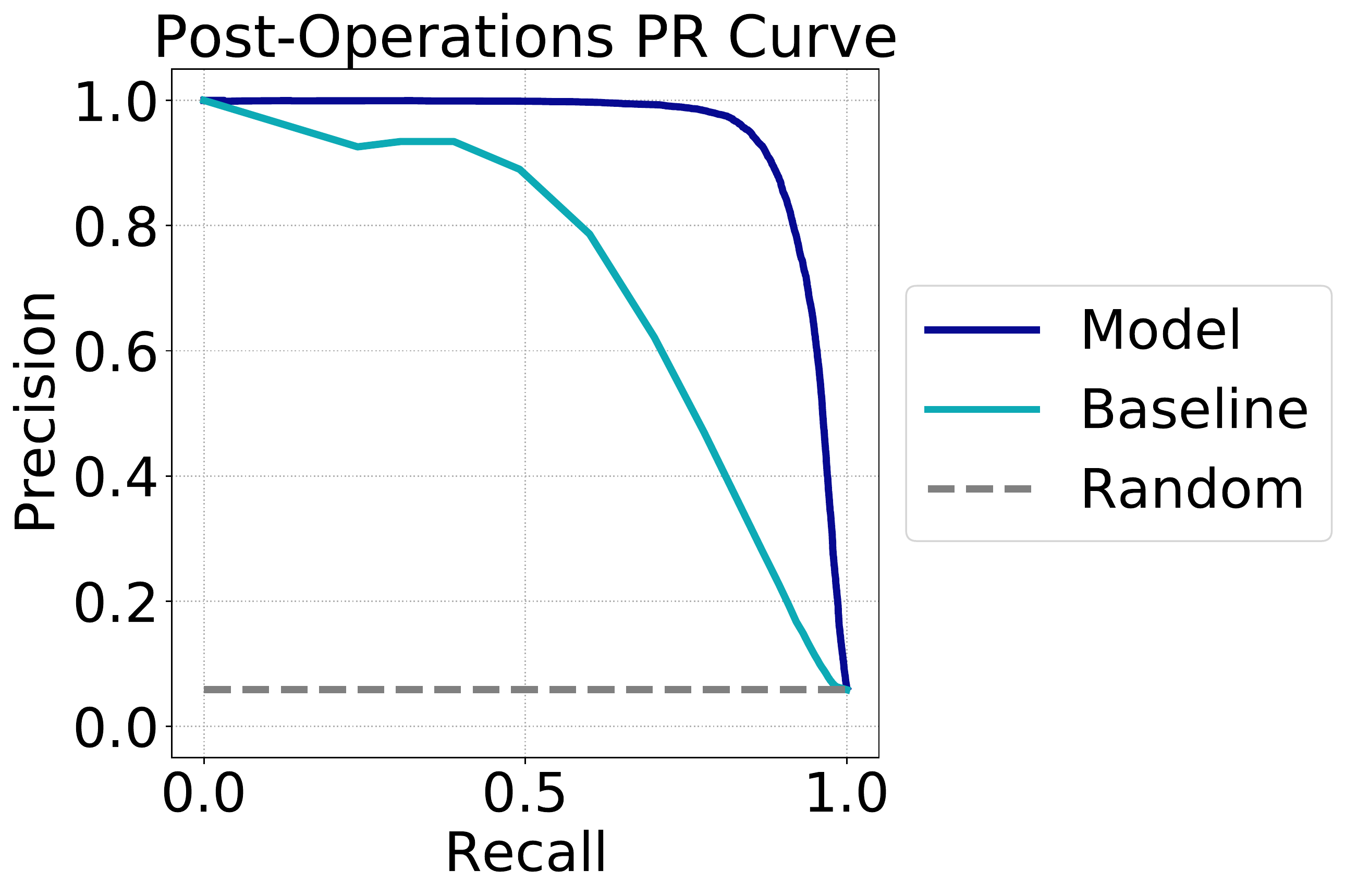}\label{fig:PostOperationsPR}}
    \caption{\ac{PR} curves for the proposed models and the corresponding baselines for the test set. Proposed models always offer better performance in terms of operating curve and precision and recall curve. The baseline classifier of Strategic and Pre-Tactical models yields an AUC$_\text{PR}$ of 0.26 while the corresponding proposed models present AUC$_\text{PR}$ of 0.90 and 0.89, respectively. Concerning the Tactical DSM, the obtained values of AUC$_\text{PR}$ were 0.70 for the baseline solution and 0.94 for the DSM. Lastly, the Post-Operations baseline gives an AUC$_\text{PR}$ equal to 0.72 and with the proposed model we obtain 0.95.
     }
    \label{fig:ResultsPR}
\end{figure*}

Training is done with 90\% of the data and the presented results are obtained with the remaining 10\%. This split is stratified to preserve the same proportions of samples of each class. Since the used data was collected throughout a single year, it would not be possible to generalize the annual seasonality in case the split was done in months. Thus, a random split is applied. 

To define the \ac{GMM} which will be used, it is necessary to select the covariance type and the number of components in the model. To do so, different information criteria were used, namely the Akaike information criterion (AIC) and the Bayesian information criterion (BIC) defined as:

\begin{equation}
\begin{split}
    AIC=-2\log { L } +2K, \\
    BIC=-2\log { L } +2K\log { N }.
\end{split}
\end{equation}
where $L$ is the likelihood, $K$ is the number of estimated parameters and $N$ is the number of samples. In both AIC and BIC, the lowest value, the best the model is.
The configuration that resulted in the best scores was the diagonal covariance type with 200 components (for all models). 
The parameters used for XGBoost are learning rate equal to 0.4 and maximum depth of 15. 

\subsection{Strategic}

The baseline model of the Strategic \ac{DSM} uses the scheduled connection time as feature since it is the information available at this stage. 

The baseline's \ac{ROC} curve presents an AUC$_\text{ROC}$ equal to 0.82 and the best threshold value is given by the optimal trade-off between \ac{FPR} and \ac{TPR}:

\begin{equation}
    FPR=\frac { FP }{ FP+TN }, \: \: TPR=\frac { TP }{ TP+FN }.
\end{equation}
where the \ac{TP}, \ac{FP}, \ac{TN} and \ac{FN} are the correct and incorrect predicted values.  

As G-mean is a metric for imbalanced classification that measures the balance between sensitivity (\ac{TPR}) and specifiticity (1-\ac{FPR}), it was the metric used to select the optimal threshold value of the baseline \ac{ROC} curve presented in Figure \ref{fig:StrategicROC}. The best threshold corresponds to 100 minutes with a G-mean value equal to 0.76.

For a \ac{PR} curve the interest is to obtain a threshold that results in the best precision and recall trade-off. This is equivalent to optimize the $\text{F}_1$ score which represents the harmonic mean of both metrics. The greater $\text{F}_1$ score of the \ac{PR} curve of the baseline classifier corresponds to the threshold of 70 minutes with a $\text{F}_1$ score value of 0.32. The AUC$_\text{PR}$ obtained with this classifier is 0.26.

Regarding the current connecting time value used by TAP, 60 minutes, the G-mean value is equal to 0.51 and the $\text{F}_1$ score value is 0.27. This means that the 60 minutes threshold yields good results when compared with the observed range, although it is not the optimal value.

With the proposed Strategic \ac{DSM}, it was obtained a G-mean score equal to 0.93 and a $\text{F}_1$ score of 0.81. This model yields and AUC$_\text{ROC}$ of 0.98 and an AUC$_\text{PR}$ equal to 0.90. Figures \ref{fig:StrategicROC} and \ref{fig:StrategicPR} show the \ac{ROC} and \ac{PR} curves, respectively, of both the baseline and the Strategic model. Comparing the results, it is possible to conclude that there was a significant improvement in classification performance. In fact, these results indicate that there are other factors that influence the passenger's ability to make the connection successfully. 

To investigate the main drivers of the model, a cooperative game-theoretic approach is used. 
SHAP~\cite{Lundberg2017APredictions}, originated in the Shapley values concept from economics~\cite{Shapley1951NotesGame}, and is an important application of the concept to ML, namely by developing efficient implementations for a few common model families and by speeding up computations via local explanations.

To give global post-hoc explanations of the model, TreeSHAP is used. TreeSHAP is a variant of SHAP for tree-based machine learning models such as XGBoost~\cite{Lundberg2020FromTrees}.

\begin{figure}[h!]
    \centering
    \includegraphics[width=\linewidth]{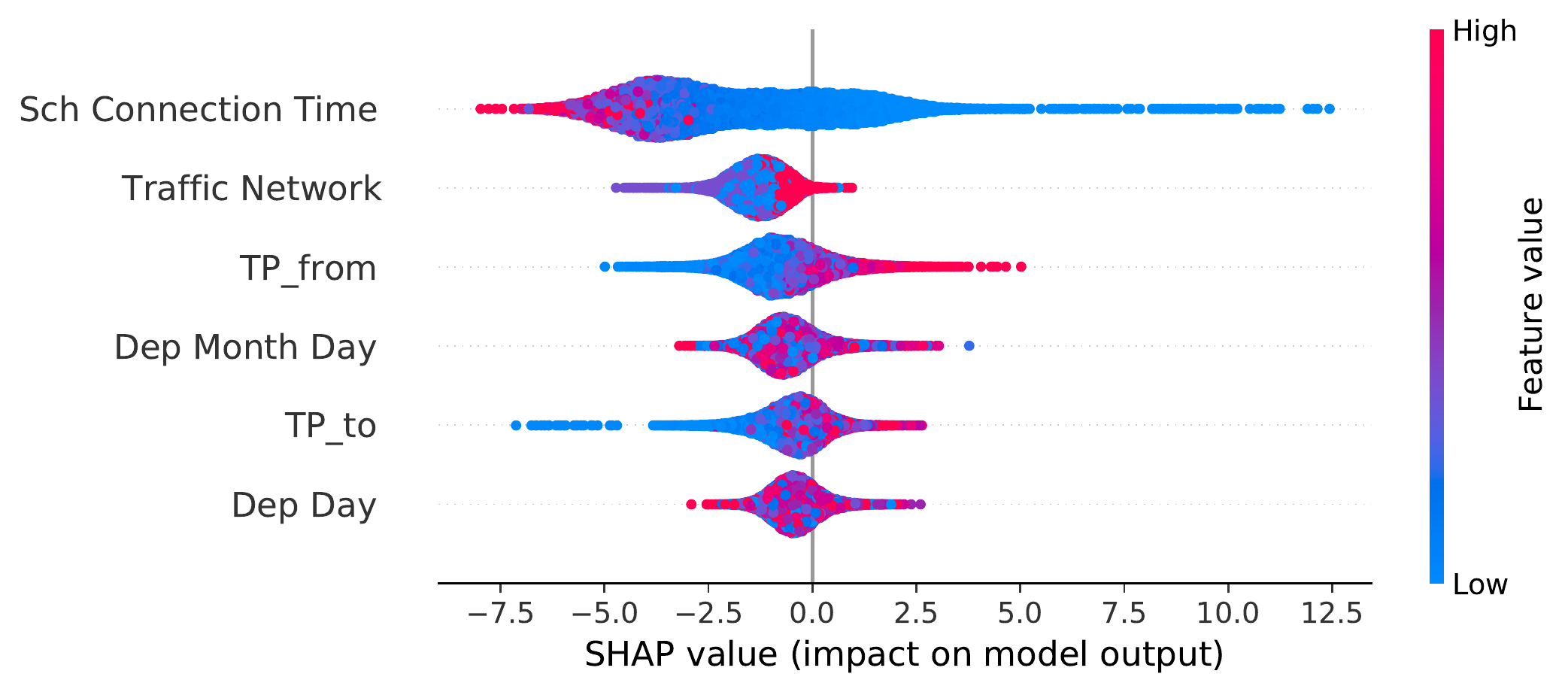}
    \caption{SHAP summary plot of the Strategic model. As expected, scheduled connection time has the largest impact on missing a flight, and the second most relevant feature relates with border control procedures. Each point in the plot corresponds to a Shapley value of an instance and a feature. The x-axis represents the Shapley value and the position along the y-axis is determined by the feature. More relevant features at the top. The importance is defined as the mean absolute value of the SHAP values for each feature. The colormap represents the feature value in a scale from low to high. For each feature, there are some jittered points along the y-axis giving a representation of overlapping points. There are no passenger features in Strategic planning.}
    \label{fig:StrategicSHAP}
\end{figure}

In Figure \ref{fig:StrategicSHAP}, we see that the feature with greatest importance value is the scheduled connection time. Samples with low connection time have positive SHAP values. This means that short connection time is likely to contribute to a positive prediction or, in other words, to classify the connections as missed. This observation is expected, since it is consistent with the already used criterion by TAP. It can also be observed that there is a high number of samples whose connection time value is more likely to be mapped to a negative prediction. Even so, there are a considerable number of samples with SHAP value between 0 and 3, approximately. This means that in fact, the time connection initially scheduled can be improved to prevent passengers from missing their connecting flights. Since this feature is the one that defines the baseline, the remaining features are responsible for improving the model's performance. 

The second feature with greatest importance value is the passenger traffic network within the airport. This highlights the fact that it is important to take into account the airport's bottlenecks (passport and security control) which the passenger has to pass through. After applying the encoding, the different categories of traffic network are numerically ordered as follows: \ac{NN} $<$ \ac{SN} $<$ \ac{SS} $<$ \ac{NS}. It is interesting to notice that the model can capture that the SS connections (in purple) are the ones with the most negative impact in the model, which means that the lack of bottlenecks makes the samples more likely to be classified as not missed connections. The model is also able to perceive that passengers travelling in a NS connection (in pink), corresponding to the traffic network with the greatest number of bottlenecks, are more likely to be classified as missed connections. 

Concerning the arrival and departure flight designators (`TP from' and `TP to', respectively), high values correspond to categories that are often associated with positive targets, hence the encoding produces the expected results. Both features present a large number of samples whose impact on model output is approximately zero. The remaining samples suggest that sometimes the origin or destination might have an influence on the result. Such influence may be related to predictable flight delays, if some flights have a historically tendency to be delayed. 

Unlike what happens with the other features, the departure week and month days, do not present a clear interpretation in terms of SHAP values. This does not mean that these features are not relevant, since the SHAP values are not always zero, but that the influence may be related to other factors in a way that is not noticeable.

\subsection{Pre-Tactical}
Comparatively to the Strategic \ac{DSM}, the new introduced information is the one relative to the passengers. That is, no information regarding flights delays was added. Therefore, the feature used in the Pre-Tactical baseline model is the scheduled connection time. All the results obtained with the baseline of the Strategic model are still valid and will be used to establish a point of comparison to the Pre-Tactical model's results.

Regarding the Pre-Tactical model, Figure \ref{fig:PreTacticalROC} presents the performance of the proposed model in \ac{ROC} space, with an AUC$_\text{ROC}$ equal to 0.98. This model yields a G-mean value of 0.92. Figure \ref{fig:PreTacticalPR} shows the obtained \ac{PR} curves for both the baseline and the Pre-Tactical \ac{DSM}. The proposed model gives an AUC$_\text{PR}$ of 0.89 and a $\text{F}_1$ score value of 0.80. From these results it is possible to conclude that the proposed solution significantly outperforms the baseline solution.

Comparing the results of the two models, Strategic and Pre-Tactical, it is interesting to note that, although new information was introduced, the results are similar. This may mean that the new features introduced regarding the passenger do not add information relevant to the predictions.

\begin{figure}[h!]
    \centering
    \includegraphics[width=\linewidth]{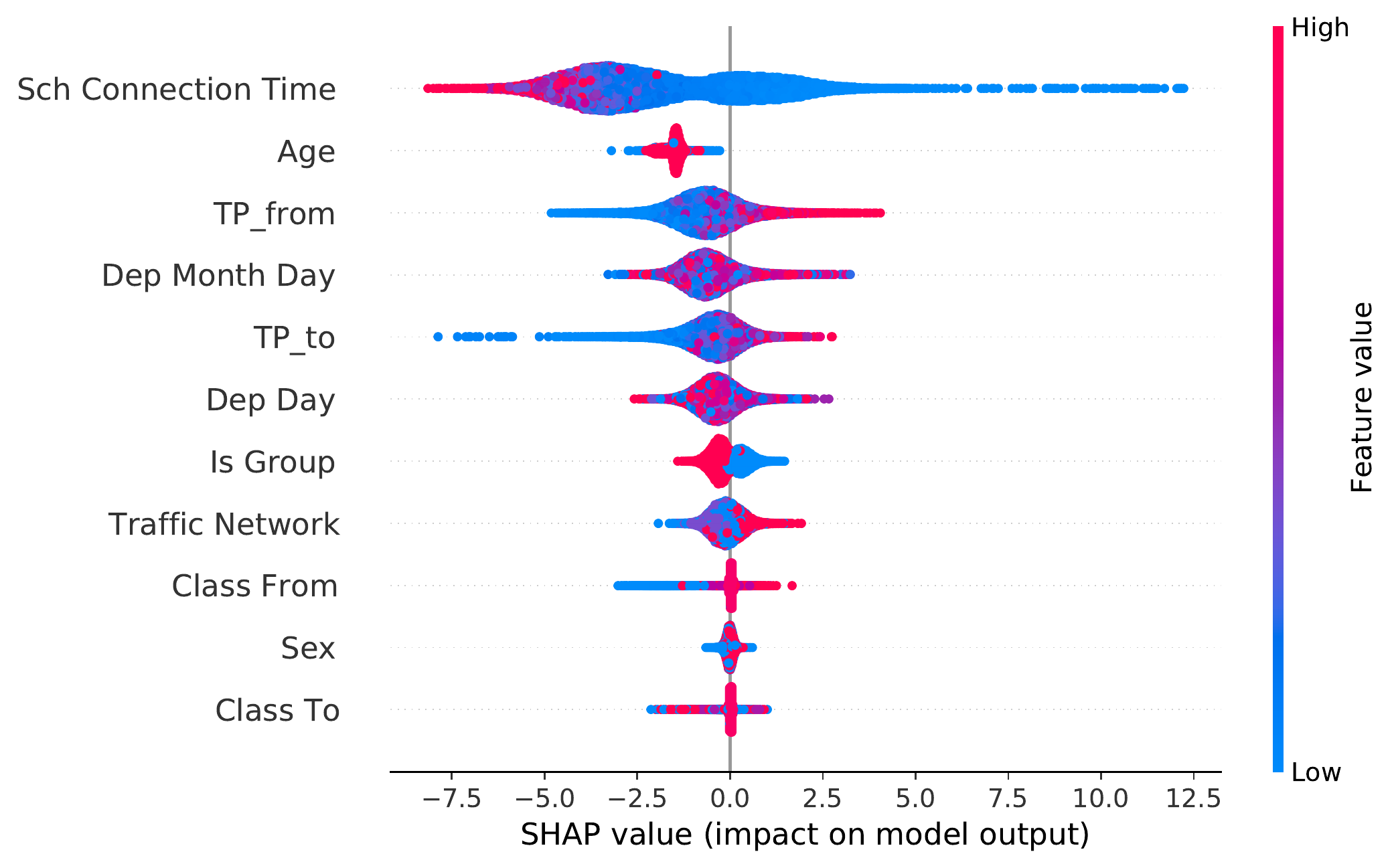}
    \caption{SHAP summary plot of the Pre-Tactical model. At this stage, passenger information is available for the predictive model. We see that passenger age is now the second most important feature.} 
    \label{fig:PreTacticalSHAP}
\end{figure}

It can be observed that the age is the feature with the second greatest feature importance value. However, as seen in the results, the passenger's features do not contribute to improve the model's performance. This can be explained by the range of SHAP values: despite having an absolute mean distant from zero, the value varies little from instance to instance. This means that the contribution of the feature is often similar, whereas a feature whose contribution varies a lot with its value is more informative.

Of the new features introduced, it is worth highlighting the behaviour of the binary variable which indicates whether the passenger travels in a group (represented in pink) or not (in blue). As can be seen from Figure \ref{fig:PreTacticalSHAP}, there is a clear division between the two classes near the SHAP value zero. Samples in blue have positive values, meaning that passengers travelling alone are more likely to be predicted as missed connecting passengers. 

Analysing the feature which represents the gender of the passenger, it can be concluded that the model cannot distinguish any trends regarding the influence of this feature.

\subsection{Tactical}
The feature used as baseline for the Tactical \ac{DSM} was the perceived connection time. Regarding the \ac{ROC} space, the AUC$_\text{ROC}$ is 0.93 and the threshold which gives the greatest G-mean value is 60 minutes. In fact, this value is the same as the one used by TAP \ref{fig:TacticalROC}. This point corresponds to a G-mean value of 0.88. Figure \ref{fig:TacticalPR} shows the performance of the baseline model in the \ac{PR} space. The optimal threshold is 70 minutes with a $\text{F}_1$ score value of 0.67. The obtained value of AUC$_\text{PR}$ was 0.70. The overall performance of this model is better than the Strategic and Pre-Tactical baseline. This may be and indicator that taking into account the arrival flight delays significantly improves the model performance.

The proposed Tactical \ac{DSM} yields an AUC$_\text{ROC}$ of 0.99 and the an AUC$_\text{PR}$ value equal to 0.94. The G-mean and $\text{F}_1$ scores of the Tactical \ac{DSM} are 0.94 and 0.88, respectively. As expected, the results obtained with the proposed model are significantly better than those obtained with the baseline. It can be concluded that this model has a better performance than the previous ones, which is in line with the intuition that the predictive capacity increases as the more information becomes available.

\begin{figure}[h!]
    \centering
    \includegraphics[width=\linewidth]{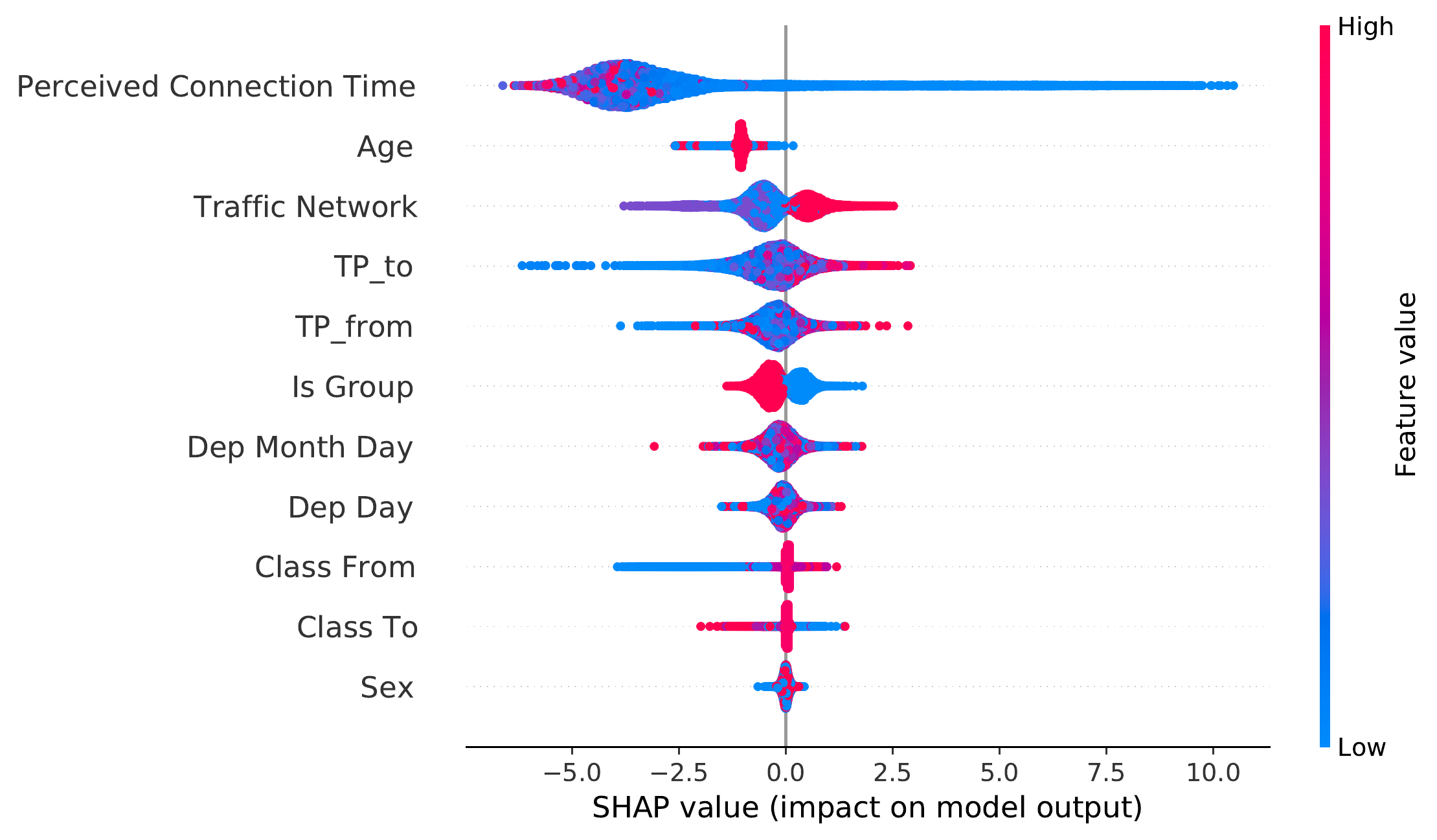}
    \caption{SHAP summary plot of the Tactical model. Our new feature, perceived connection time, shows high importance over all others. Now passenger age is followed by traffic network, that is more discriminative for missing a flight, when bringing in the passenger perception of how much time she will have until departure of her connection.}
    \label{fig:TacticalSHAP}
\end{figure}

From the SHAP summary plot presented in Figure \ref{fig:TacticalSHAP} it is possible to analyse that the overall behaviour of the features is the same as in the previous models. Note that concerning the traffic network, the division of the feature classes is more evident and there is a clear separation between the \ac{NS} connections (in pink) and the other type of connections. In fact, \ac{NS} samples are the only ones with positive SHAP values. 

Regarding the perceived connection time, it is interesting to note that, unlike the scheduled connection time, most of the samples have negative SHAP values. This means that the majority of the samples are more likely to be mapped to the negative class, i.e., not missed connections.

\subsection{Post-Operations}

The Post-Operations DSM uses the actual connection time as the feature for the baseline model. In the \ac{ROC} space, the best threshold is 70 minutes with a G-mean value of 0.87, whereas in the \ac{PR} space the optimal threshold is 40 minutes for a $\text{F}_1$ equal to 0.68. In fact, the metrics' results are similar to the ones from the Tactical baseline. This is an indicator that the departure flight delay (new information added) does not improve the performance of the model. A possible reason may be that most departure delays occur after boarding ends, so passengers have already boarded and are no longer on the verge of missing their flight. The obtained values of AUC$_\text{ROC}$ and AUC$_\text{PR}$ were 0.93 and 0.72, respectively. 

The Post-Operations model gives an AUC$_\text{ROC}$ of 0.99 (Figure \ref{fig:PostOperationsROC}) and an AUC$_\text{PR}$ of 0.95 (Figure \ref{fig:PostOperationsPR}). The G-mean and $\text{F}_1$ score values obtained are 0.94 and 0.89, respectively. As can be seen, the model significantly outperforms the baseline solutions. However, when compared to the Tactical \ac{DSM} results, the metrics' values are almost the same. This indicates that, contrary to what was expected, knowing the actual connection time does not improve the classification performance.

\begin{figure}[h!]
    \centering
    \includegraphics[width=\linewidth]{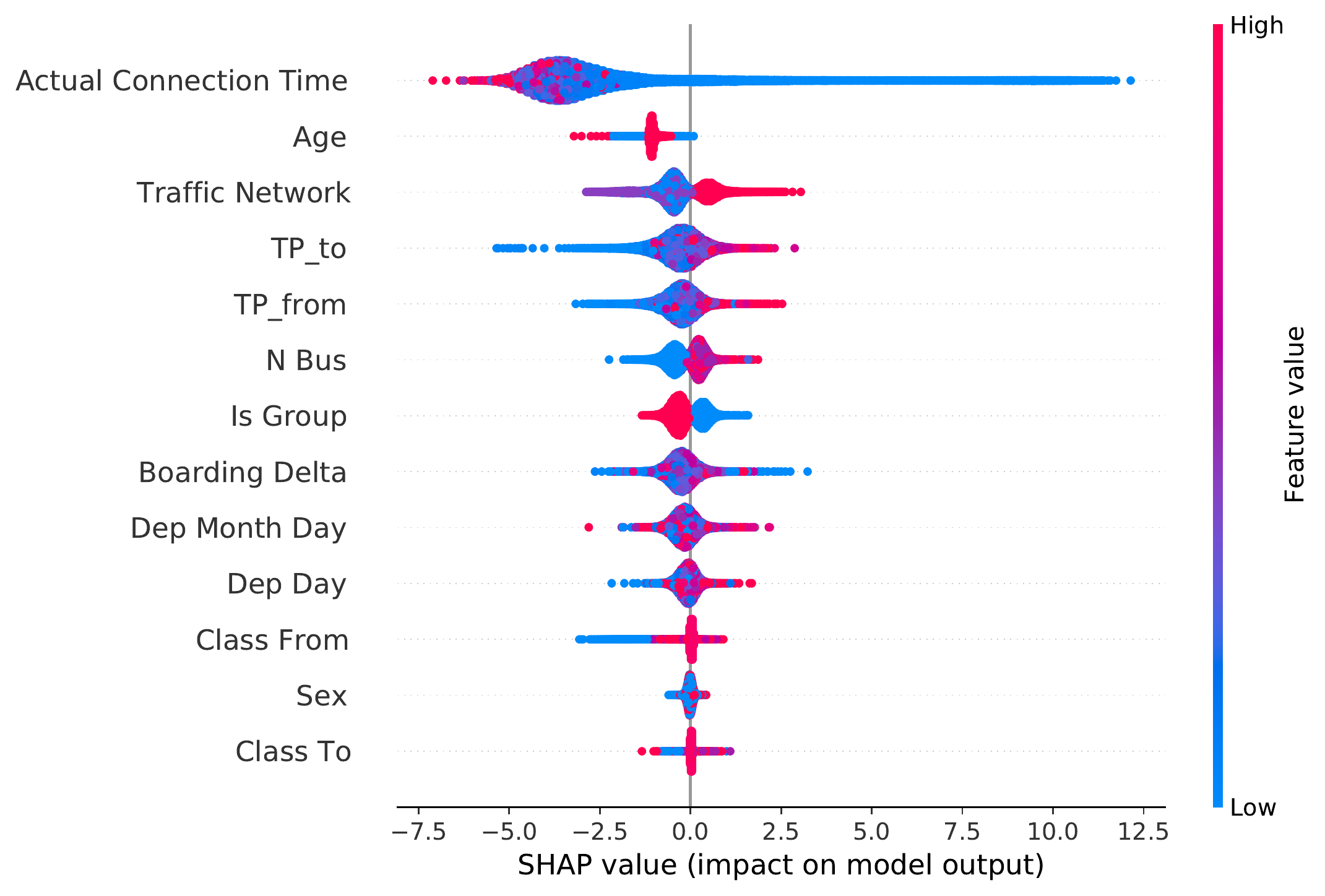}
    \caption{SHAP summary plot of the Post-Operations model. In the results of the analysis model, we input the true connection time, so to assess the main drivers of the flight loss behavior. Again, the contribution of age is indecisive and points management to further investigate this variable, through more data collection and elicitation of domain knowledge. Important factors to consider when scheduling have been, unsurprisingly, if there are border controls needed (Traffic Network), origin, destination and if buses are used. To be considered in other operational aspects like price brokering, if the trip is a group trip.}
    \label{fig:Pos-OperationsSHAP}
\end{figure}

As can be seen from Figure \ref{fig:Pos-OperationsSHAP}, the SHAP values of the actual connection time are very similar to the ones of the perceived connection time from the Tactical model (Figure \ref{fig:TacticalSHAP}). In fact, this is an indicator that the departure flight delay (new information added) does not improve the performance of the model. A possible reason may be that most departure delays occur after boarding ends, so passengers have already boarded and are no longer on the verge of missing their flight.

It is interesting to note that the model distinguishes a clear division between samples whose value for the number of buses is zero (in blue) or greater than zero (in shades of purple and pink). A possible explanation for this division may be the fact that when no buses are used, the boarding is done by the jet bridge. Therefore, the boarding gates remain open until closer to the actual departure time and passengers have more time to catch the flight. On the other hand, when buses are used to take passengers from the boarding gate to the platform where the plane is located, the boarding time is possibly reduced to ensure that the rest of the processes take place on schedule. Thus, samples representing the later case have positive impact on the model output meaning that the samples are more likely to be predicted as missed connections. It is also possible to analyse that in the cases where buses are used, the number is not relevant. In other words, there does not seem to be a relationship between the number of buses (and consequently the number of passengers to be transported) and the respective impact on the model's output. This actually makes sense: since passengers are already being transported to the plane, they will not lose their connection.

Regarding the feature boarding delta, its value does not have a clear relationship with its impact on the model output. 

\subsection{Operational Costs}
\label{sec:operational-costs}
The results presented and discussed in the previous Sections show that for the same model different results are obtained according to the used metric. To choose the best result it is necessary to do a trade-off between  \ac{TP}, \ac{FP}, \ac{TN} and \ac{FN}, that is, a trade-off between precision and recall. A possible solution that helps to identify which of these metrics deserves more attention is to consider the relative cost of preventive and reactive actions in each model. Reaction refers to the actions that have to be taken when a connection is missed. These include allocating a new departing flight or pay for the overnight stay. Prevention includes any action taken in advance to prevent passengers from missing the connection. These actions can range from allocating an employee to guide passengers during the connection, to delaying the departure flight. Although the costs of these actions can be quite varied, average costs $C_{REAC}$ and $C_{PREV}$ are considered. In this work, cases in which preventive actions are not successful and, hence, imply reactive actions, are not considered. 
Table \ref{tab:Costs} shows how the number of reactive and preventive actions can be obtained from the confusion matrix.
\begin{table}[!htb]
\caption{Number of reactive and preventive actions.}
  \renewcommand{\arraystretch}{1.2} 
  \centering
  \begin{tabular}{lcc}
    \toprule
    & No prevention &   Model based prevention \\
    \midrule
    \# Reactions & TP+FN & FN \\
    \# Preventions & --  & TP+FP \\

    \bottomrule
  \end{tabular}
  \label{tab:Costs}
\end{table}
The operational costs variation is given by the difference between the cost of the model based preventions and taking no preventive actions. That is,

\begin{equation}
\begin{split}
    \Delta C=\left[C_{REAC}\cdot FN+C_{PREV}\cdot(TP+FP)\right]\\ -C_{REAC}\cdot(TP+FN).
\end{split}
\end{equation}

Assuming that the ratio between reactive and preventive costs is given by $C_{REAC}=r\cdot C_{PREV}$, to make our study dimensionless,
\begin{equation}
   \Delta C= C_{PREV}\cdot(TP+FP)-r\cdot C_{PREV}\cdot TP,
\end{equation}
which means that to have a negative variation (cost reduction), it is necessary to verify
\begin{equation}
\begin{split}
    C_{PREV}\cdot\left[TP+FP-r\cdot TP\right]<0 \Leftrightarrow   r>\frac { TP+FP }{ TP }\\ \Leftrightarrow r>\frac { 1 }{ precision }.
\end{split}
\end{equation}

\begin{table}[!htb]
\caption{Ratio between reactive and preventive costs for each model.}
  \renewcommand{\arraystretch}{1.2} 
  \centering
  \begin{tabular}{lcc}
    \toprule
    & Precision &   $r_{min}$ \\
    \midrule
    Strategic & 0.73 & 1.37 \\
    Pre-Tactical & 0.73 & 1.37 \\
    Tactical & 0.86  & 1.16 \\

    \bottomrule
  \end{tabular}
  \label{tab:CostsRatio}
\end{table}

The proposed \acp{DSM} solutions, as presented in Table \ref{tab:CostsRatio}, are cheaper than the model with no prevention, when $r>r_{min}$ (i.e., these solutions represent a cost reduction if the cost of taking a reactive action is at least $r_{min}$ greater than the preventive solution). The obtained $r_{min}$ values are considered to be small, which means that, in general, the use of the proposed models easily pays off.

From Table \ref{tab:CostsRatio} it is possible to conclude that the ratio between reactive and preventive actions can be lower in the Tactical model while still representing a cost reductions. This is consistent with the fact that preventive actions closer to departure time have a prevention cost closer to the reaction cost.

Since the Post-Operations \ac{DSM} is applied as an analytical model after all events have happened, it does not allow preventive actions and, therefore, a similar analysis is not possible to perform.

The minimum ratio between reactions and preventives cost so that the model pays off depends only on precision. However, recall defines how much the model actually pays off (in terms of absolute range). The optimal trade-off between precision and recall could be obtained by having the effective values of $C_{REAC}$ and $C_{PREV}$.

\section{Conclusions and Discussion}

Our models can predict with high precision whether a given passenger can make the connection successfully or not, by framing the learning task as a classification problem.

The introduction of a new feature, the scheduled connection time discounted by the arrival delay, for the tactical, real time model allowed for an increase of predictive performance and changed feature importance scores attributed by SHAP.

The need to retrain the models periodically due to the seasonality and the evolution of the aeronautical sector is a limitation of the method. Nevertheless, it is a common limitation in predictive modeling approaches. A future line of work would be to study the out-of-distribution behavior of the models, and to adapt an online learning methodology.  

We developed \acp{DSM} that allow better planning of an airline connecting flights in a European Airport. These models can be used in the different stages of operations: Strategic (long-term), Pre-Tactical (short-term), Tactical (real-time) and Post-Operations analysis.
With an AUC of the ROC over 0.98, these models can help the decision-maker take preventive actions and easily represent a reduction of operational costs.
Additionally, the prevention of missed connections improves passenger experience and, thus, satisfaction and loyalty towards the airline. Lastly, this work identifies some key factors that potentially impact passengers’ connection times.


%



\section*{Acknowledgment}

The authors would like to thank TAP Air Portugal, for providing the data, and Eng. Duarte Afonso, for sharing all the domain knowledge.

\ifCLASSOPTIONcaptionsoff
  \newpage
\fi



%

%

\end{document}